\RequirePackage{fix-cm}
\documentclass[twocolumn]{svjour3}          
\smartqed  
\usepackage{amsmath}
\usepackage{nccmath}
\usepackage{amsfonts}
\usepackage{graphicx}
\usepackage[numbers,sort,compress]{natbib}
\begin{document}

\title{Data-driven identification of nonlinear dynamical systems with LSTM autoencoders and Normalizing Flows
}


\author{Abdolvahhab Rostamijavanani \and Shanwu Li         \and
        Yongchao Yang 
}


\institute{Abdolvahhab Rostamijavanani (Corresponding Author) \at
              Department of Mechanical Engineering-Engineering Mechanics, Michigan Technological University, Houghton, Michigan 49931, USA\\
              \email{arostami@mtu.edu}    
}


\maketitle

\begin{abstract}
While linear systems have been useful in solving problems across different fields, the need for improved performance and efficiency has prompted them to operate in nonlinear modes. As a result, nonlinear models are now essential for the design and control of these systems. However, identifying a nonlinear system is more complicated than identifying a linear one. Therefore, modeling and identifying nonlinear systems are crucial for the design, manufacturing, and testing of complex systems. This study presents using advanced nonlinear methods based on deep learning for system identification. Two deep neural network models, LSTM autoencoder and Normalizing Flows, are explored for their potential to extract temporal features from time series data and relate them to system parameters, respectively. The presented framework offers a nonlinear approach to system identification, enabling it to handle complex systems. As case studies, we consider Duffing and Lorenz systems, as well as fluid flows such as flows over a cylinder and the 2-D lid-driven cavity problem. The results indicate that the presented framework is capable of capturing features and effectively relating them to system parameters, satisfying the identification requirements of nonlinear systems.

\keywords{Nonlinear dynamics  \and Dynamical systems \and System Identification  \and  Deep learning  \and Normalizing Flows}
\end{abstract}

\section{Introduction}\label{sec:1}

 Linear systems have been successfully used to solve various problems across different fields. However, the demand for better performance and efficiency pushes these systems to operate in nonlinear modes, which necessitates the use of nonlinear models for their design and control. Identifying a nonlinear system is a much more intricate process than identifying a linear one. Therefore, it is essential to model and identify nonlinear systems for designing, manufacturing, and testing complex systems. The models for nonlinear systems can be broadly categorized into White Box~\cite{liu2021dendrite,giordano2018black,ljung2010perspectives,picard2016comparison}, Black Box~\cite{polifke2014black,gretton2001support,sjoberg1995nonlinear,kim2012subspace,ayala2020nonlinear}, and Grey Box~\cite{kim2012subspace,pires2023nonlinear,vizer2013local,mohammadi2015new,beneventi2012effective} modeling to identify the system~\cite{schoukens2019nonlinear}. White Box modeling is effective but challenging as it requires accurate physics knowledge and is difficult to generalize to various systems. Black Box modeling is an alternative to physics-based modeling and characterizes the model based on experimental data, making it easy to use but less interpretable. 
 
 Black Box modeling deals with estimating a function by hypothesizing a functional relationship between input and output, making it simpler but difficult to estimate and interpret the parameters.  Experiment design becomes more challenging, model selection requires more careful consideration, and parameter estimation is more complex. Therefore, it is important to go beyond traditional linear methods of identification and move towards advanced nonlinear methods of identification. Ideally, a nonlinear system identification approach should be recommended when the function preprocessing step indicates that the distortion levels are excessively high and significantly surpass the data noise floor \cite{schoukens2019nonlinear}. 

The Black Box model can be solved using several techniques, both classical~\cite{petsounis2001parametric} and modern~\cite{juditsky1995nonlinear}, such as neural networks. The use of data-driven approaches has become increasingly popular in modeling nonlinear systems due to the advancements in machine learning and deep learning technology. The fundamental architecture of deep learning, known as deep neural networks (DNNs), has proven to be particularly effective in capturing the intrinsic features of complex systems. DNNs offer a remarkable modeling capacity and learning flexibility, allowing them to represent complex relationships in a hierarchical manner. The universal approximation theorem ~\cite{hornik1990universal,Akhtar2018} is a significant result in the field of deep learning. It states that a DNN with an adequate number of neural units and nonlinear activations can represent a wide range of intricate functions.  For example, DNNs have the capability to learn and represent complex nonlinear relationships between input and output variables. DNNs also provide the flexibility of adapting the network architecture to different tasks. This means that the architecture of the DNN can be tailored to the specific characteristics of the problem at hand, such as the identification of nonlinear dynamics. This adaptive design of the network architecture enables DNNs to achieve high accuracy in complex modeling tasks, making them a powerful tool in the field of nonlinear systems. LSTM autoencoder and Normalizing Flows are two deep learning models that can be used for nonlinear system identification, each with its unique features.

LSTM autoencoder~\cite{lazzara2022surrogate,simpson2021machine,xu2019multitask} can be used to extract temporal features of dynamical systems from time series data, which can then be used for further analysis or modeling. This approach can be useful in various applications such as signal processing, system identification, and anomaly detection. Normalizing Flows~\cite{urain2020imitationflow,kobyzev2020normalizing,khader2021learning,sun2022probabilistic}  is a class of generative models that aims to learn the underlying probability distribution of a given dataset. By using Normalizing Flows, it is possible to capture the underlying probability distribution of the data and use it to generate new data samples or infer the latent variables that explain the observed data. By combining these two models~(creating a black-box), we can extract meaningful temporal features from the time series data using LSTM autoencoder and use Normalizing Flows to map these features to system parameters. This can facilitate nonlinear system identification in various applications such as control, prediction, and modeling of complex systems

Therefore, the primary objective of this study is to create a deep learning framework capable of identifying nonlinear systems. The framework is designed to address two main challenges. The first challenge is the extraction of features from the time series data (i.e., response data) of the dynamical system using a LSTM-encoder. The second challenge involves relating these features to system parameters using a Normalizing Flows framework.To put it differently, the black box consists of two models. The first model is responsible for capturing time-related features of the data (e.g., signal frequency), while the second model establishes a relationship between these features and system parameters. By utilizing deep learning, this process guarantees nonlinear transformations, ensuring that the system identification is based on a nonlinear approach.
\section{Methodology: Normalizing Flows concept}
Normalizing Flows is a generative model~\cite{urain2020imitationflow,kobyzev2020normalizing,khader2021learning,sun2022probabilistic} that efficiently and accurately estimates density by producing tractable distributions. In contrast, GANs~\cite{goodfellow2020generative,ogunmolu2016nonlinear,gedon2021deep} and VAEs~\cite{kingma2019introduction,ma2020structural,yan2021deep} do not explicitly learn the probability function of training data. Instead, GANs generate similar data to deceive the discriminator in a min-max game until it reaches a saturation point where it cannot differentiate between real and fake samples. VAEs, on the other hand, learn to identify variational inference in latent space and generate data using a decoder. However, neither GANs nor VAEs can learn the real probability density functions (PDFs) of real data. Normalizing Flows, a rigorous generative method, learns the real PDF of a dataset by using invertible and differentiable functions. To create a random variable $\mathcal{X}$ with a complicated distribution $\mathcal{P}$, Normalizing Flows applies an invertible and differentiable function ($\textit{f}$) to a random variable $\mathcal{Z}$ with a simple distribution, such as a standard normal distribution $\mathcal{Z} \sim N(0, 1)$ using this formula: $\mathcal{ X} = f(\mathcal{Z}) \sim \mathcal{P}$ and $\mathcal{Z}=f^{-1}(\mathcal{X})$. The change of variables method can be used to calculate the transferred distribution $\mathcal{P}$ as follows:
\begin{equation}
    log \hspace{0.1cm}\mathcal{P}(\mathcal{X})= log \hspace{0.1cm}\mathcal{P}(\mathcal{Z})-log\hspace{0.1cm}|\frac{\partial f}{\partial \mathcal{Z}}(\mathcal{Z})|
\end{equation}
However, finding a single function (bijector) that transfers the distribution in the desired manner is not an easy task. In cases where the target distribution $\mathcal{P}$ is highly complex, a simple $\textit{f}$ (e.g., a scale or shift function) is not adequate. To address this issue, we can create a more intricate PDF by composing bijectors with one another, forming a more complex chain of bijectors. The following example demonstrates this approach:
\begin{equation}\label{eq2}
    \textit{f}=\textit{f}_{k}\hspace{0.1cm} \circ \hspace{0.1cm}f_{k-1}\hspace{0.1cm} \circ\hspace{0.1cm}...f_{1}
\end{equation}
A Normalizing Flows involves transforming a base distribution (such as the standard normal distribution) into a more complex target distribution by applying a sequence of bijectors one after another:
\begin{equation}
\begin{split}
   \mathcal{Z}_0&=\mathcal{Z}\\
   \mathcal{Z}_{k}&=f_{k}(\mathcal{Z}_{k-1})\\
   \mathcal{X}&=\mathcal{Z}_{k}
\end{split}
\end{equation}
In addition, the transformed (target) distribution can be obtained by summing the contributions from each bijector:
\begin{equation}
  log \hspace{0.1cm}\mathcal{P}(\mathcal{X})= log \hspace{0.1cm}\mathcal{P}(\mathcal{Z})-\sum_{i=1}^{i=k}log\hspace{0.1cm}|\frac{\partial f_{i}}{\partial \mathcal{Z}_{i-1}}(\mathcal{Z}_{i-1})|
\end{equation}
To obtain the target distribution, we can assign each $\textit{f}_i$ simple functions such as scale and shift, along with a basic nonlinearity such as sigmoid or ReLU function. It's worth noting that each $\textit{f}_i$ has parameters (like scale and shift values) that can be learned from training data using \textit{maximum likelihood} estimation.
\subsection{Models with Normalizing Flows}
All Normalizing Flows models share the properties of being both invertible and differentiable. The RealNVP model (Real-valued Non-Volume Preserving)\cite{dinh2016density} can be constructed by stacking a sequence of bijectors that have easily computable Jacobian determinants and are easily invertible. An earlier alternative to RealNVP is Non-linear Independent Component Estimation (NICE)\cite{dinh2014nice}. Autoregressive models are another type of Normalizing Flows that have fast-to-compute Jacobian matrices because each $\textit{f}i$ (see equation. \eqref{eq2}) only depends on $\mathcal{Z}{1},...,\mathcal{Z}i$. Therefore, ${\frac{\partial f{i}}{\partial \mathcal{Z}{j}}=0}$ whenever $\mathrm{j>i}$, resulting in a lower triangular Jacobian matrix, with the determinant being a simple product of the diagonal elements. Additionally, the joint density $\mathcal{P}(\mathcal{X})$ can be modeled as the product of conditionals $\prod{i} \mathcal{P}(\mathcal{X}{i}|\mathcal{X}{1:i-1})$. In this paper, we utilize a masked autoregressive flow model.
\begin{figure*}[htbp]
	\centering
	\includegraphics[width=0.9\textwidth]{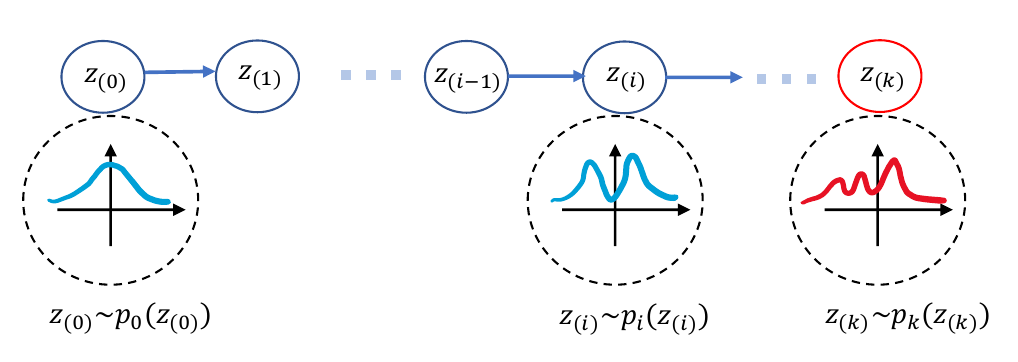}
	\caption{The concept of Normalizing Flows can be defined as the process of transforming a simple probability distribution function ($\textit{P}_0$) to the original, more complex distribution function ($P_k$) by utilizing bijectors. }
	\label{NF1}
\end{figure*}
\begin{figure*}[htbp]
	\centering
	\includegraphics[width=0.9\textwidth]{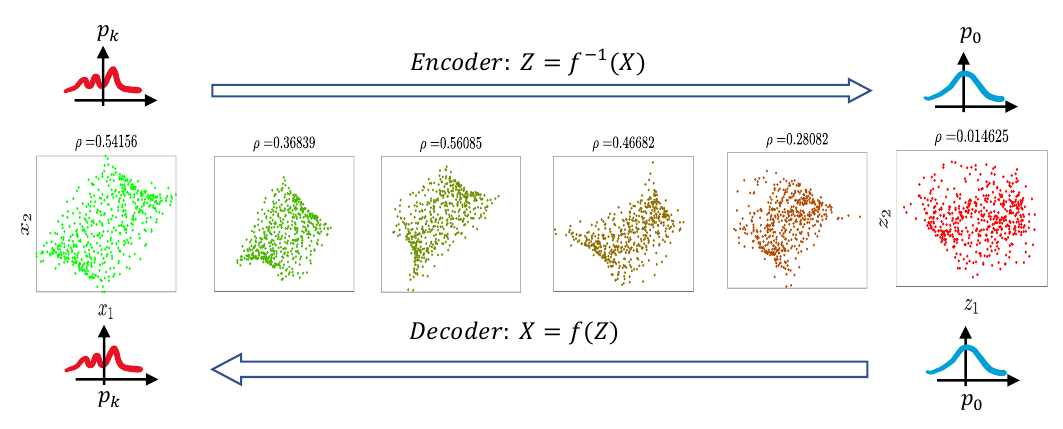}
	\caption{In the Normalizing Flows concept, the dependency between the two original coordinates of a 2-DOF Duffing system can be reduced by passing through Normalizing Flows layers. Here, $Z_0$ represents the decomposed modal coordinates, and $X$ refers to the original coordinates.  }
	\label{NF2}
\end{figure*}
\begin{figure*}[htbp]
	\centering
	\includegraphics[width=0.90\textwidth]{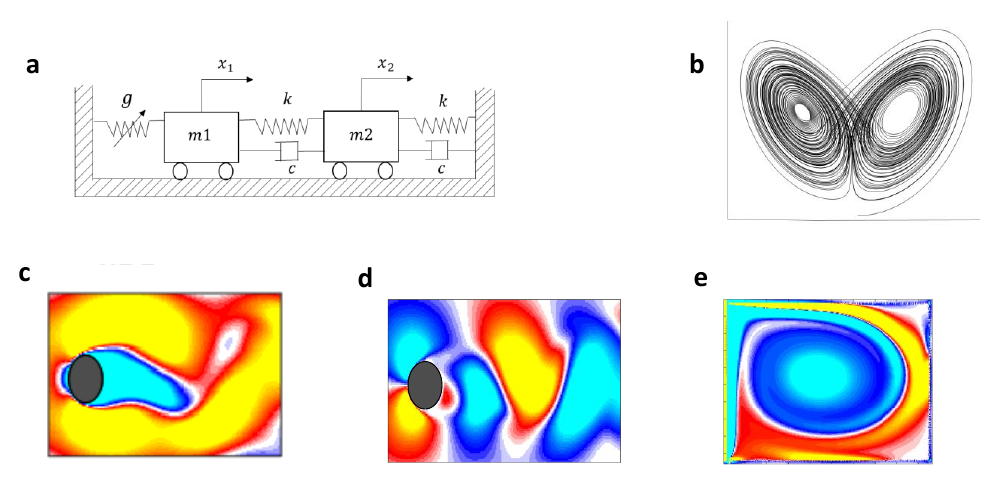}
	\caption{Case studies: \textbf{a}: 2 DOF Duffing system \textbf{b}: Lorenz system \textbf{c}: Stream-wise velocity over a cylinder \textbf{d}: Transverse velocity over a cylinder \textbf{e}: Vorticity of a 2-D driven-lid cavityproblem }
	\label{examples}
\end{figure*}
\begin{figure*}[htbp]
	\centering
	\includegraphics[width=0.92\textwidth]{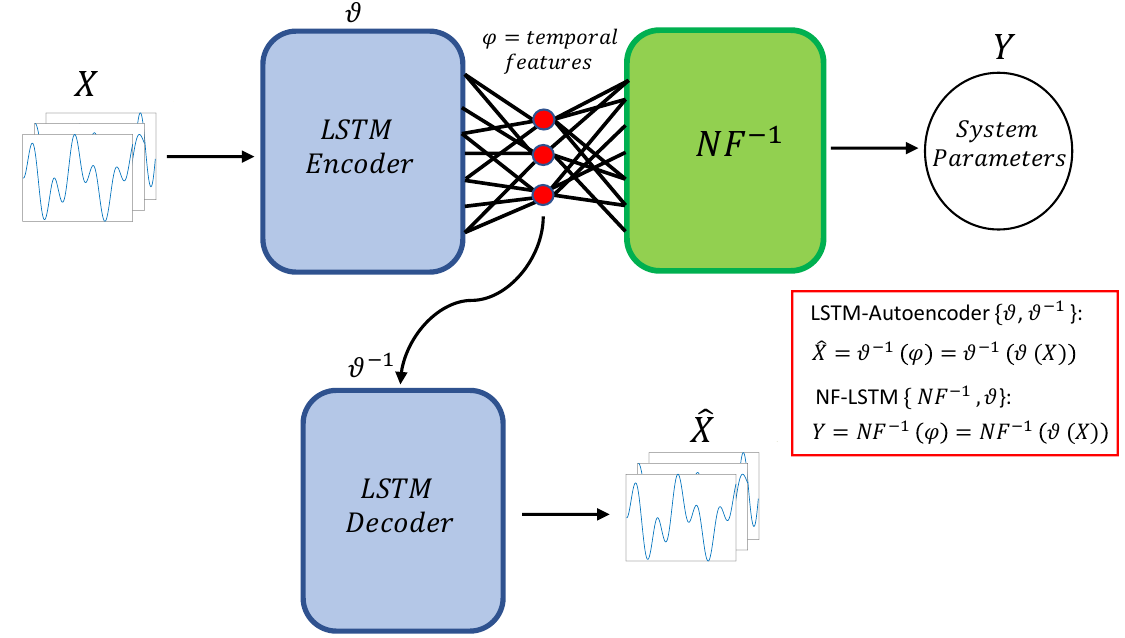}
	\caption{The architecture of the framework presented for non-fluid case studies involves two main components. Firstly, the temporal features~($\varphi$) of the input data (i.e., trajectories: $X$) are extracted using LSTM-Encoder ($\vartheta$). Next, Normalizing Flows (NF) is utilized to establish a relationship between these temporal features and the system parameters~($Y$). In order to ensure that the temporal features of the input data represent the most critical information, a LSTM-Decoder ($\vartheta^{-1}$) is utilized to reconstruct the data.   }
	\label{SI_1}
\end{figure*}
\begin{figure*}[htbp]
	\centering
	\includegraphics[width=0.92\textwidth]{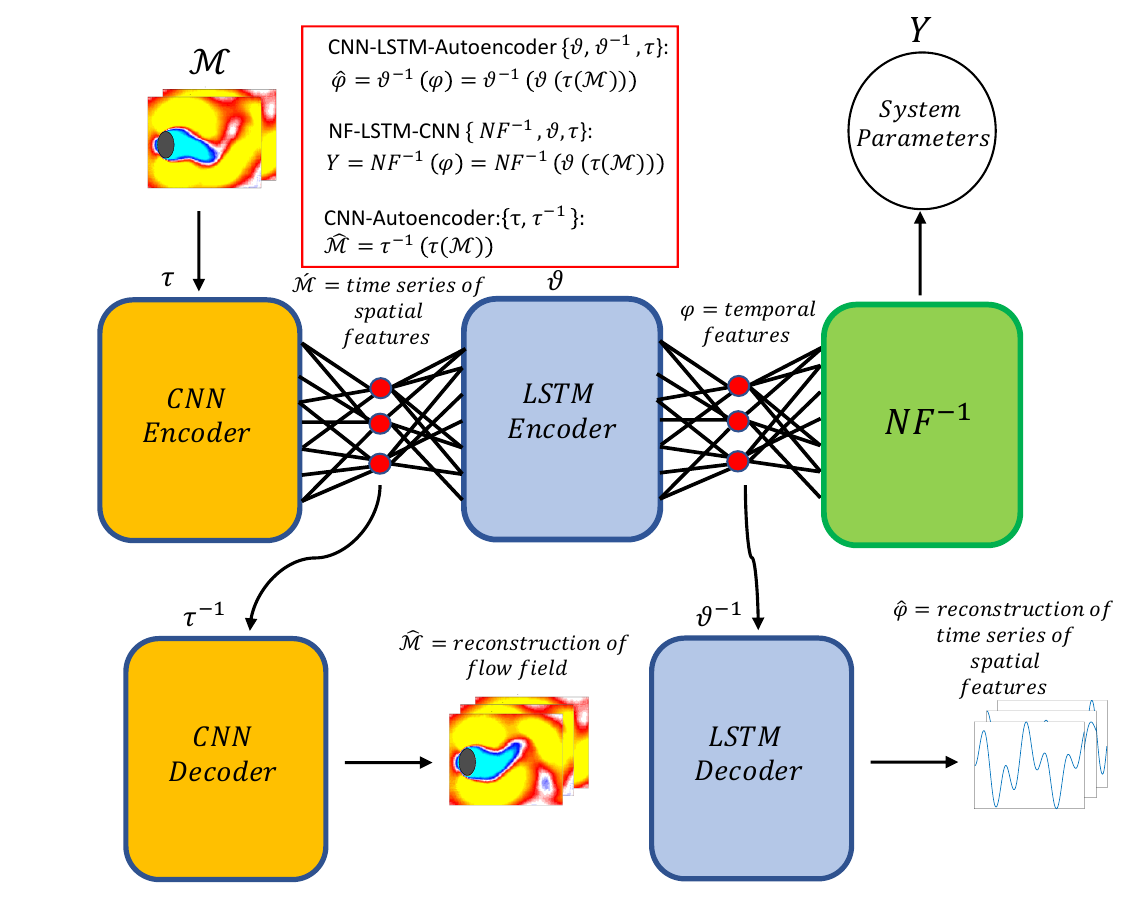}
	\caption{The architecture of the framework presented for fluid case studies involves three main components. Firstly, the spatial features ($\mathcal{M}^\prime$) of the flow field ($\mathcal{M}$) is extracted using a CNN-Encoder ($\tau$) .Secondly, the temporal features~($\varphi$) of the spatial features data are extracted using LSTM-Encoder. Next, Normalizing Flows (NF) is utilized to establish a relationship between these temporal features and the system parameters~($Y$). In order to ensure that the spatial and temporal features of the input data represent the most critical information, a CNN-Decoder ($\tau^{-1}$) and a LSTM-Decoder ($\vartheta^{-1}$) are used respectively to reconstruct the data.  }
	\label{SI_2}
\end{figure*}
\begin{figure*}[htbp]
	\centering
	\includegraphics[width=0.95\textwidth]{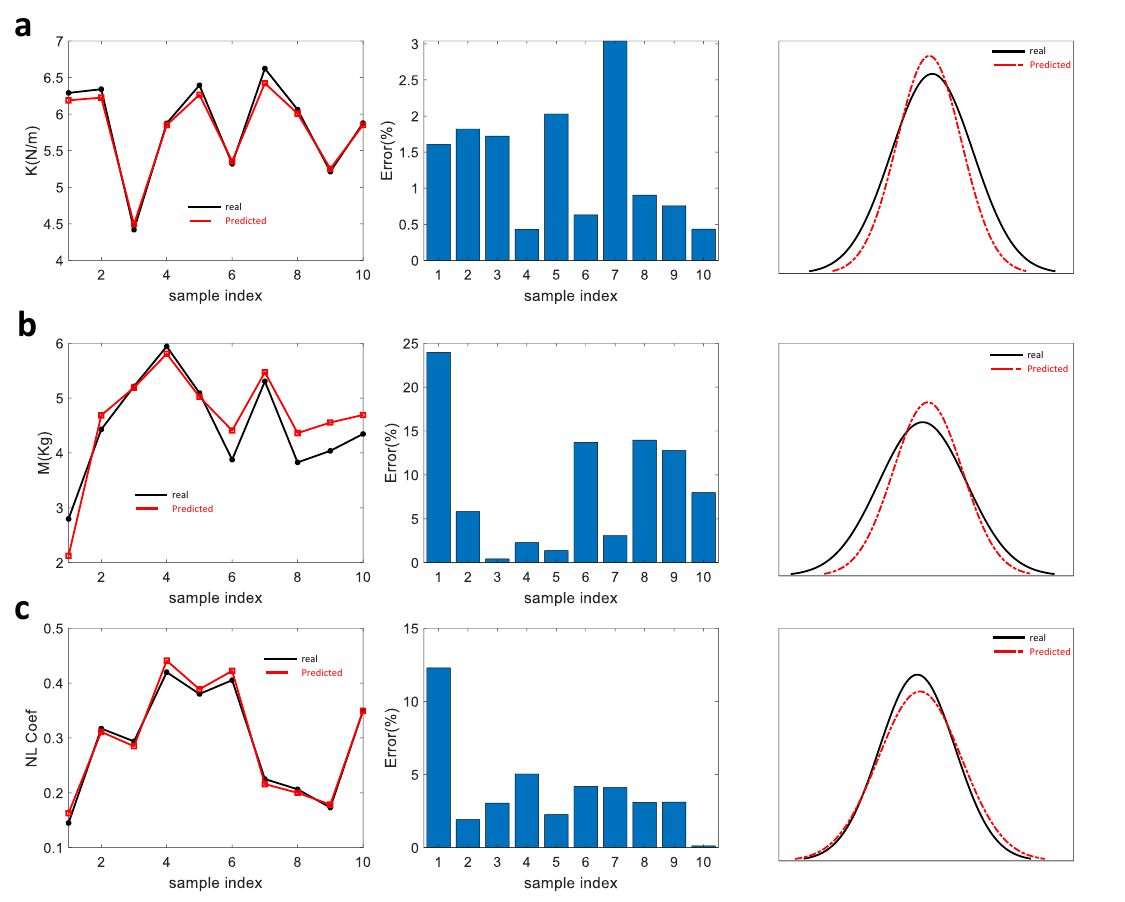}
	\caption{system parameters identification of 2-D Duffing systems. \textbf{a} Only the stiffness (K) of system changes over samples. Comparison of the parameters predicted by the framework and the ground truths and the distribution of predicted and ground truths parameters. \textbf{b} Only the mass (M) of system changes over samples. \textbf{c} Only the nonlinearity (NL Coef) of system changes over samples.    }
	\label{duffing}
\end{figure*}

\begin{figure*}[htbp]
	\centering
	\includegraphics[width=0.95\textwidth]{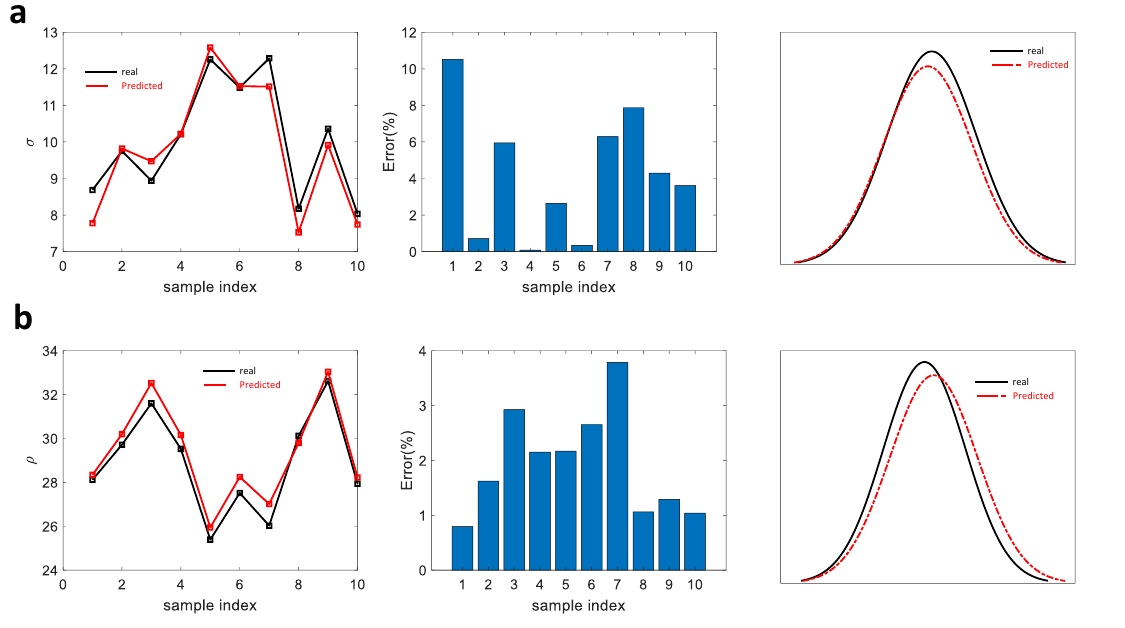}
	\caption{system parameters identification of Lorenz systems. \textbf{a} Only parameter $\sigma$ of system changes over samples. Comparison of the parameters predicted by the framework and the ground truths and the distribution of predicted and ground truths parameters. \textbf{b} Only parameter $\rho$ of system changes over samples.   }
	\label{lor1}
\end{figure*}

\begin{figure*}[htbp]
	\centering
	\includegraphics[width=0.95\textwidth]{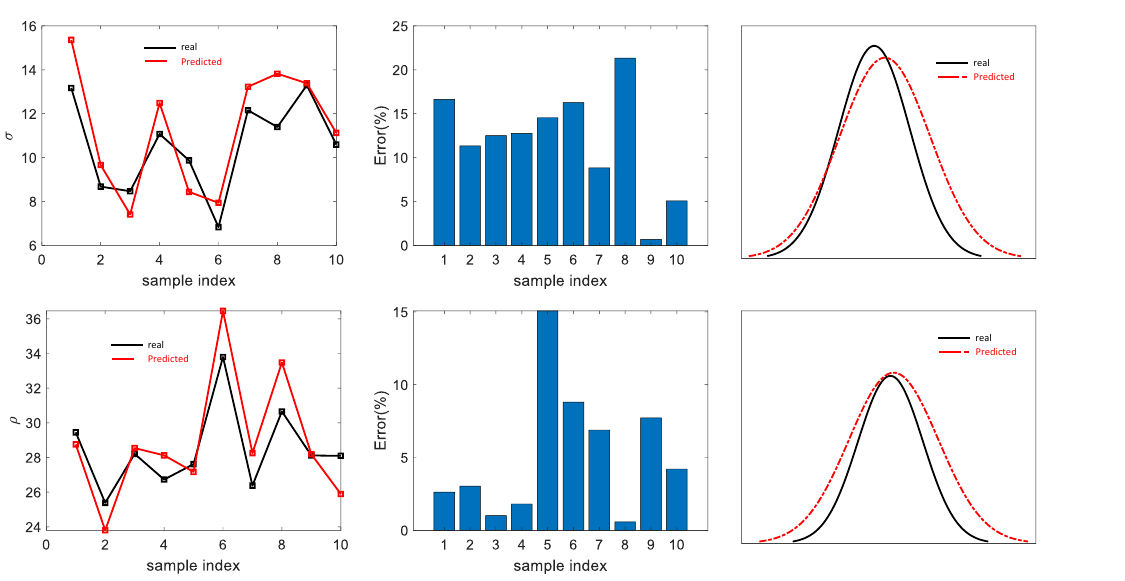}
	\caption{system parameters identification of Lorenz systems. Two system parameters ($\sigma$ and $\rho$) change together over samples. Comparison of the parameters predicted by the framework and the ground truths and the distribution of predicted and ground truths parameters show the performance decreases as the situation is more complex for the framework compared to single-parameter identification.  }
	\label{lor2}
\end{figure*}

\section{Problem formulation}\label{sec:2}
System identification is a mathematical approach that aims to uncover the fundamental aspects of a system that control its behavior. This method is particularly useful in solving inverse problems where we know the output of a system, such as displacement or velocity trajectories of a Duffing system or the velocity or pressure field of a fluid flow system, but we want to determine the system's parameters, denoted by $P_{sys}$. Given the input-output data of a nonlinear system, the goal of nonlinear system identification is to determine the system's underlying dynamics and parameters, which can be used to predict the system's future behavior. In order to achieve this objective, we utilize a function denoted as $F$, which establishes the relationship between the output of the system, $\mathcal{R} \in R^n$, and the parameters that govern its behavior, $P_{sys}\in R^s$ where $n$ and $s$ are the number of degrees of freedom of the system and number of corresponding system parameters, respectively. The formulation is as beloow:
\begin{equation} \label{eq1}
    P_{sys}=F(\mathcal{R})
\end{equation}
 However, this relationship can be complex, and different sets of parameters may produce similar system behavior, making it challenging to determine the correct parameters. Therefore, we also employ another function, $\mathcal{K}$, which connects some of the system's features~($f$), such as its frequency, to the parameters. 
\begin{equation}
    P_{sys}=\mathcal{K}(f)
\end{equation}
while $f$ is achieved directly form system output ($\mathcal{R}$) by the function $G$ as bellow:
\begin{equation}
    f=G(\mathcal{R}) \label{eq2}
\end{equation}
As for fluid flow systems, due to their high-dimensionality, we first attempt to capture the spatial features before extracting the temporal features and subsequently the system parameters. To achieve this, we use $\mathcal{R}^\prime\in R^m$~($m<n$) as follows:
\begin{equation}
    \mathcal{R}^\prime=\tau(\mathcal{R})
\end{equation}
where $\tau$ is a function that captures the spatial features. We are able to utilize $\mathcal{R}^\prime$ in place of $\mathcal{R}$ in equations \eqref{eq1} and \eqref{eq2} for identifying the system parameters.

\begin{figure*}
	\centering
	\includegraphics[width=0.95\textwidth]{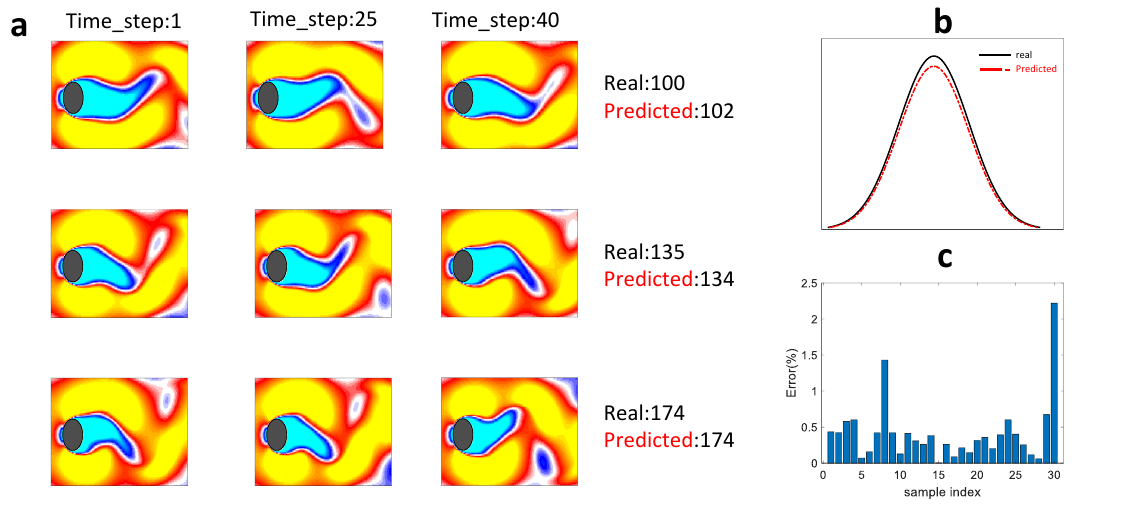}
	\caption{system parameters identification of flow over a cylinder (stream-wise velocity). \textbf{a} Three Reynolds numbers identified by framework using a sequence of flow fields. \textbf{b} Comparison of the parameters predicted by the framework and the ground truths and the distribution of predicted and ground truths parameters. \textbf{c} The error between the identified system parameters and the ground truth values.  }
	\label{vel_u}
\end{figure*}

\begin{figure*}
	\centering
	\includegraphics[width=0.95\textwidth]{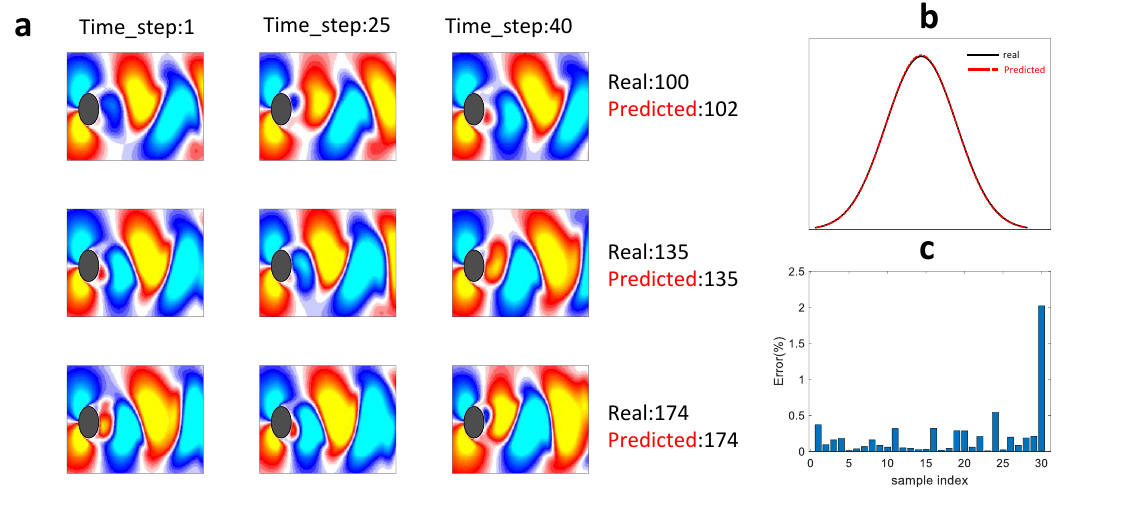}
	\caption{system parameters identification of flow over a cylinder (transverse velocity). \textbf{a} Three Reynolds numbers identified by framework using a sequence of flow fields. \textbf{b} Comparison of the parameters predicted by the framework and the ground truths and the distribution of predicted and ground truths parameters. \textbf{c} The error between the identified system parameters and the ground truth values.    }
	\label{vel_v}
\end{figure*}

\section{Deep learning framework}\label{sec:3}
\subsection{Objective}
Due to the lack of a general mathematical framework for nonlinear system identification, the primary goal of this approach is to identify the parameters of a given dynamical system. Specifically, by analyzing time-series data obtained from the Duffing system, we aim to identify important parameters such as the mass, stiffness, and nonlinearity of the system. Similarly, in fluid flow data, we can determine the Reynolds number by analyzing two-dimensional flow fields over time. This framework is capable of capturing unique features of dynamical systems and extracting their corresponding parameters.
\begin{figure*}[htbp]
	\centering
	\includegraphics[width=0.95\textwidth]{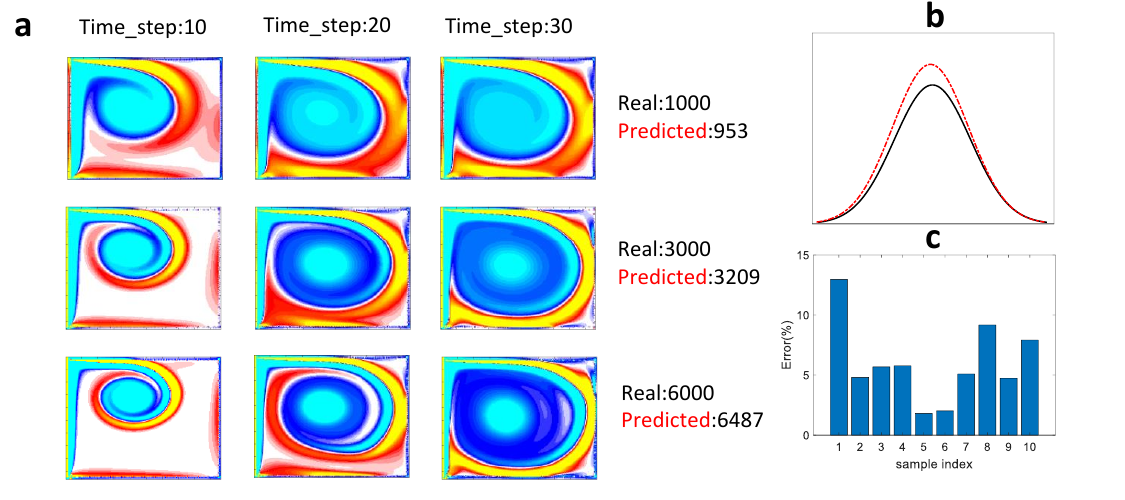}
	\caption{system parameters identification of vorticity of a 2-D lid-driven cavity problem. \textbf{a} Three Reynolds numbers identified by framework using a sequence of flow fields. \textbf{b} Comparison of the parameters predicted by the framework and the ground truths and the distribution of predicted and ground truths parameters. \textbf{c} The error between the identified system parameters and the ground truth values.    }
	\label{cavity}
\end{figure*}
\subsection{Deep Normalizing flow framework}
\subsubsection{2-DOF Duffing oscillator and Lorenz Systems}
One of the examples presented in this study is the 2-DOF Duffing oscillator. Duffing systems are composed of three parameters: mass, stiffness, and nonlinearity coefficients~($M,K,N\!L~ coef$). Each study modifies one of these parameters while holding the other two constant. Another case study involves the Lorenz systems, which are chaotic dynamical systems controlled by three parameters ({ $\sigma, \beta, \rho$ }). To extract the primary features of the data, such as frequency, the time series data is encoded using an LSTM-Encoder and subsequently decoded to return to the original data. Once the features are extracted, they are passed to the Normalizing flow to identify the system parameters. \par
The framework presented captures the temporal characteristics of the system and establishes a connection with system parameters. The overall loss function is defined as follows:
\begin{equation}
    \mathcal{L}= {\alpha _{N\!F}}{\mathcal{L}_{N\!F}} + {\alpha _{rec-lstm}}{\mathcal{L}_{rec-lstm}}+ {\alpha _{rec-f}}{\mathcal{L}_{rec-f}}
\end{equation}
Loss functions used in in this case are expressed as below:
\begin{enumerate}
\itemsep=0pt
\item $\mathcal{L}_{NF}$: The loss function for temporal features~($\varphi$) reconstruction is negative log-likelihood (NLL). The first loss function utilized is probability density estimation through Normalizing flow. The loss function is as follows:
\begin{equation}
   \mathrm{\mathcal{L}_{\mathcal{N}}=-\frac{1}{ns}\sum_{i=1}^{i=ns}log\hspace{0.1cm} \mathcal{P}(\varphi^{(i)})}
\end{equation}
 where $\varphi$ is the features extracted from LSTM-autoencoder, $(i)$ denotes the index number of training dataset and $ns$ represents the number of training samples.

\item $\mathcal{L}_{rec-lstm}$: The time series data is reconstructed using an LSTM-Autoencoder. The LSTM-encoder acts as a forward nonlinear transformation that converts the original time series data to temporal features in the latent intrinsic space. The decoder subsequently transforms the feature coordinates back to their original coordinates. The associated loss is trajectory reconstruction, which force the LSTM-Autoencoder to reconstruct the original time series data from the latent space. The corresponding loss is therefore minimizing:
\begin{equation}
 \frac{1}{ns}\sum_{i=1}^{i=ns}||X-\vartheta^{-1}\left(\vartheta\left( X\right)\right)||^{(i)}_{{\text{MSE}}}   
\end{equation}
where $\mathrm{ns}$ represents the number of training samples and $\mathrm{(i)}$ denotes the index number assigned to each sample.
\item $\mathcal{L}_{rec-f}$: Auxiliary feature reconstruction. In order to establish a connection between system parameters and temporal features, we provide the corresponding system parameters of each sample as input to the NF model and connect them to the temporal features using the following loss function: 
\begin{equation}
 \frac{1}{ns}\sum_{i=1}^{i=ns}||\varphi-N\!F(Y))||^{(i)}_{{\text{MSE}}}   
\end{equation}
where $Y$ is the corresponding system parameters of $X$ data and $\varphi$ is the corresponding temporal features extracted by the LSTM-Encoder~(see Fig. \ref{SI_1})
\end{enumerate}
\begin{figure*}[htbp]
	\centering
	\includegraphics[width=0.95\textwidth]{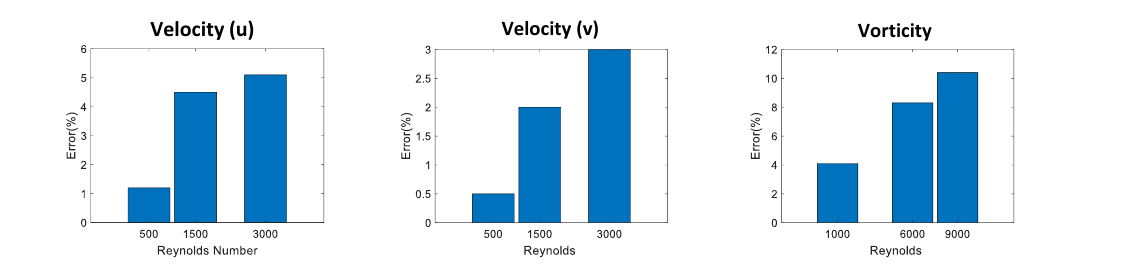}
	\caption{The effect of Reynolds number on the performance of the presented framework for three case studies.   }
	\label{reynolds}
\end{figure*}


\subsubsection{Flow passing a cylinder and 2-D cavity problem}
In this section, we present another example where we analyze a 2-D flow over a cylinder and a 2-D cavity problem with a driven lid. Initially, the 2-D spatial time series data~($\mathcal{M}$) is inputted into a CNN-Encoder~($\tau$), which reduces the spatial size and extracts the most significant spatial features/patterns of flow~($\mathcal{M}^\prime$). The spatial features are then sent to a LSTM-encoder~($\vartheta$) to extract the temporal features~($\varphi$) from the spatial features. Additionally, the spatial features are processed by a CNN-decoder~($\tau^{-1}$) to restore them to the original 2-D time series flow. The extracted temporal features are then sent to the NF model to map to the flow parameter~($Y$), Reynolds number. Furthermore, the LSTM-Decoders~($\vartheta^{-1}$) are employed to convert the extracted temporal features to spatial features. It is worth noting that the CNN-Decoder and LSTM-Decoder assist in determining whether the extracted spatial/temporal features are reliable.\par
The overall loss function is defined as follows:
\begin{equation}
    \mathcal{L}= {\alpha _{N\!F}}{\mathcal{L}_{N\!F}} + {\alpha _{rec-lstm}}{\mathcal{L}_{rec-lstm}}+ {\alpha _{rec-CNN}}{\mathcal{L}_{rec-CNN}}+ {\alpha _{rec-f}}{\mathcal{L}_{rec-f}}
\end{equation}
The regarding loss functions are as:
\begin{enumerate}
 \item Reconstruction of parameters loss function: The negative log-likelihood (NLL) loss function is utilized for Normalizing flow probability density estimation~(similar to the approach we used for the Duffing and Lorenz systems). The loss function is defined as below:
\begin{equation}
   \mathrm{\mathcal{L}_{\mathcal{N}}=-\frac{1}{ns}\sum_{i=1}^{i=ns}log\hspace{0.1cm} \mathcal{P}(\varphi^{(i)})}
\end{equation}
 where $\varphi$ is the features extracted from LSTM-autoencoder and $(i)$ denotes the index number of training dataset.
    \item $\mathcal{L}_{rec-CNN}$: Reconstruction of time series 2-D flow by CNN-Autoencoder. To decrease the dimensionality of the flow data, a CNN-Autoencoder is constructed. In order to guarantee that the extracted spatial features provide the most optimal low-dimensional representation of the flow, a reconstruction loss function is applied in the following manner:
    \begin{equation}
         \frac{1}{ns}\sum_{i=1}^{i=ns}|| \mathcal{M}-\tau^{-1}\left(\tau\left( \mathcal{M}\right)\right)||^{(i)}_{{\text{MSE}}}  
    \end{equation}
 
    \item $\mathcal{L}_{rec-lstm}$: Reconstruction of time series of spatial features by LSTM-Autoencoder. The subsequent step involves extracting the most crucial temporal features from the time series of spatial features. A reconstruction loss function is utilized to achieve this aim. The purpose of this function is to ensure that these extracted temporal features are suitable for converting back to the original time series spatial features. This is achieved by minimizing the loss function as follows: 
    \begin{equation}
          \frac{1}{ns}\sum_{i=1}^{i=ns}||\mathcal{M}^\prime-\vartheta^{-1}\left(\vartheta\left(\mathcal{M}^\prime\right)\right)||^{(i)}_{{\text{MSE}}}
    \end{equation}
  \item $\mathcal{L}_{rec-f}$: Auxiliary feature reconstruction. In order to establish a connection between system parameters~(Reynolds numbers) and temporal features, we provide the corresponding system parameters of each sample as input to the NF model and connect them to the temporal features using the following loss function: 
\begin{equation}
 \frac{1}{ns}\sum_{i=1}^{i=ns}||\varphi-N\!F(Y))||^{(i)}_{{\text{MSE}}}   
\end{equation}
where $Y$ is the corresponding system parameters of $\mathcal{M}$ data and $\varphi$ is the corresponding temporal features extracted by the LSTM-Encoder.

\end{enumerate}

\subsubsection{Network architecture and training}
\paragraph{Normalizing flow architecture:}
Autoregressive models are potent models for probability density estimation in normalizing flows. RealNVP and NICE are specific types of autoregressive models. In this study, we incorporate multiple autoregressive models into our framework. Each layer of the model comprises three dense layers, each with 512 neurons. We use nonlinear activation functions for all layers because we aim to achieve nonlinear transformation/identification. Keras, a high-level API in Tensorflow, offers various nonlinear activation functions such as Relu, Sigmoid, and tanh. Relu is among the most widely used functions due to its faster training run time \cite{lusch2018deep}, which we employ in our networks. After each autoregressive layer, we apply a permutation layer because normalizing flow layers only operate on a portion of the data, leaving the rest unchanged. Therefore, we rearrange the data spaces to ensure that all data undergo nonlinear transformation through the network (see Fig. \ref{SI_2}).
\paragraph{LSTM-Autoencoder:}
The LSTM-Autoencoder is an efficient model for extracting temporal features from time series data. We utilized this model to extract relevant features and correlated them with system parameters in this study. We applied this model in two different case studies: one involving fluid flow and another non-fluid flow. The architecture of this model, including the number of layers and neurons for each sub-model, is reported in Table \ref{tab:2}. 

\paragraph{CNN-Autoencoder:}
Some parts of the network architecture relies on a convolutional autoencoder. The convolutional layer is responsible for extracting the most critical flow field features, while the pooling layer reduces the flow's degree of freedom in the encoder block. By using these layers in the encoder, we obtain more compressed fields that contain valuable flow information. Once the latent layers capture the intrinsic flow physics, we use upsampling and convolutional layers in the decoder block to reconstruct the original flow field. To preserve the size of the reconstructed flow field, we utilize fully-connected layers, reshape layers, and flatten layers.

 We use the Adam optimizer with a slow learning rate of $\alpha=1e-5$ for both models. It is important to note that depending on the case study, all three models (for flow data) or two models (for Duffing and Lorenz systems) are trained simultaneously. This means that all weights are shared and modified during the training phase. We used the Xavier initialization method \cite{glorot2010understanding} to initialize the weights of each model. We evaluated the performance of the DNN across various training sessions by conducting hyperparameter tuning. We tested different sets of hyperparameters (weights of loss functions), and the results are based on the hyperparameters that yield the minimum validation errors (Table \ref{tab:3}).
\subsection{Datasets for training and testing}
To generate datasets for Duffing and Lorenz systems, we use the ode45 solver in MATLAB to solve the ordinary differential equations of MDOF systems, and obtain training and testing data for each scenario by randomly sampling a single parameter from a normal distribution while keeping the rest fixed. For Duffing systems where stiffness changes, we sample stiffness from a normal standard distribution with mean and standard deviation identical to those in the base distribution used in the DNN framework, and the other parameters are fixed at specific values. Similarly, for the cases where mass or nonlinearity changes, we fix the other two parameters and randomly sample the changing parameter from the corresponding normal distribution. Specifically, the training/test data for stiffness changes come from {$K\sim N(5,1)$, $M\sim N(5,0)$, $g\sim N(0.3,0)$}, the training/test data for mass changes come from {$M\sim N(5,1)$, $K\sim N(5,0)$, $g\sim N(0.3,0)$}, and the training/test data for nonlinearity changes come from {$K\sim N(5,0)$, $M\sim N(5,0)$, $g\sim N(0.3,0.1)$}. 
We follow a similar approach for the Lorenz system, where we change only one or two parameters while keeping the rest constant in order to generate different scenarios. In one scenario, we randomly sample the values of the parameters as {$\sigma\sim N(10,2)$, $\beta\sim N(8/3,0)$, $\rho\sim N(28,0)$}, while in another scenario we change the value of $\rho$ by sampling it from a normal distribution with mean 28 and standard deviation 2, and the other parameters are fixed at specific values, i.e., {$\sigma\sim N(10,0)$, $\beta\sim N(8/3,0)$, $\rho\sim N(28,2)$}. In a third scenario, we change the values of $\sigma$ and $\rho$ while keeping $\beta$ constant, i.e., {$\sigma\sim N(10,2)$, $\beta\sim N(8/3,0)$, $\rho\sim N(28,2)$}.
This process allows us to generate a range of datasets with different parameter combinations, which we can then use to train and evaluate our DNN models.\par

In the study of streamwise/transverse flow over a cylinder, we use a grid of 96 by 192 evenly spaced points to examine 100 different samples for training and 30 samples for testing, each consisting of 50 time steps. It is important to note that the Reynolds numbers are randomly sampled from a normal distribution with a mean of 150 and a standard deviation of 25. For the cavity problem, we adopt a similar approach by using a grid size of 192 by 192 and sampling Reynolds numbers from a normal distribution with a mean of 450 and a standard deviation of 25. In this case, we examine 100 different samples for training and 30 samples for testing, each containing 30 time steps.  This approach allows us to generate a range of datasets with different Reynolds numbers, which we can use to train and evaluate our models for these flow problems.

\begin{table*}
 \begin{center}
\def~{\hphantom{0}}
  \begin{tabular}{lccc}
        \\
        \hline
      Block &      $LSTM$ layer   & $MaxPooling$ layer   &   $Dense$ layer  \\[3pt]
       \hline
       CNN-Encoder   & $6~(3\times3)$  & $6~(2\times2)$  & $-$ \\
       CNN-Decoder   & $6~(3\times3)$  & $6~(2\times2)$  & $1$
  \end{tabular}
   \caption{CNN-Autoencoder architecture}
  \label{tab:1}
  \end{center}
\end{table*}

\begin{table*}
 \begin{center}
\def~{\hphantom{0}}
  \begin{tabular}{lcccc}

        \\
        \hline
      Block &      $LSTM~layer$      &   $Dense ~layer$  \\[3pt]
       \hline
       LSTM-Encoder   &3    &~~-\\
       LSTM-Decoder   &4    &1
  \end{tabular}
   \caption{LSTM-Autoencoder architecture}
  \label{tab:2}
  \end{center}
\end{table*}
\begin{table*}
 \begin{center}
\def~{\hphantom{0}}
  \begin{tabular}{lcccc}

        \\
        \hline
      Case tudy &      $\alpha_{N\!F}$   & $\alpha_{rec-lstm}$   &   $\alpha_{rec-C\!N\!N}$ &   $\alpha_{rec-f}$  \\[3pt]
       \hline
       Duffing-Lorenz   &1  & 1  &1 &1\\
       Flow over a cyllinder-Cavity problem   &1  & 1  &1 &1
  \end{tabular}
   \caption{Loss weights}
  \label{tab:3}
  \end{center}
\end{table*}
\section{Result and Discussion}
\subsection{2-DOF Duffing oscillator}
In the field of dynamic analysis, Duffing systems have been frequently used as case studies. In this section, we present a two-dimensional Duffing system as an example. The governing equation for this system is given by Eq. \eqref{eq1}:
\begin{equation}\label{eq1}
\mathbf{M}\ddot {\mathbf{x}} + \mathbf{C}\dot{\mathbf{x}} +\mathbf{K}\mathbf{x}+ \mathbf{g} = 0
\end{equation}
where $\textbf{[M]}$, $\textbf{[C]}$, and $\textbf{[K]}$ are the mass, damping, and stiffness matrices, respectively. ${\bold x}$ is the displacement vector (refer to Fig. \ref{examples}), and $\textbf{g}$ is the nonlinear term in the equation that characterizes the nonlinearity intensity. The expression for $\textbf{g}$ is given by Eq. \eqref{eq4}:
\begin{equation}\label{eq4}
\begin{split}
\mathbf{g}=\left\{\begin{array}{l} 
\mu_{1} k_{1}\delta_{1}^3-\mu_{2} k_{2}\delta_{2}^3\\
\mu_{2} k_{2}\delta_{2}^3-\mu_{3} k_{3}\delta_{3}^3\\
\end{array} \right\}
\end{split}
\qquad \qquad \qquad
\begin{split}
\left\{\begin{array}{l} 
\delta_{1}\\
\delta_{2}\\
\end{array} \right\}=\left\{\begin{array}{l} 
x_{1}\\
x_{2}-x_{1}\\
\end{array} \right\}
\end{split}
\qquad \qquad
\mathbf{\mu}=\left\{\mu_{1},\mu_{2}\right\}^T
\end{equation}
In this system, Rayleigh damping is assumed with the definition of $\textbf{[C]=q [k]}$ where $\textbf{q}$ stands for the damping coefficient.

We first test the effectiveness of our presented NF-DNN with a 2-DOF nonlinear Duffing system. The goal is to identify one of the system parameters, including the stiffness ($\textit{k}$), mass ($\textit{m}$), and nonlinearity coefficient, by first extracting features from the time series data and then using the neural network to compute the parameter. It is worth noting that only one parameter is varied, and a set of time series data is generated for training. Figure. \ref{duffing} shows the predicted parameters for three different scenarios, where each scenario involves changing only one parameter. The left plots present the predicted parameters for the test dataset (size 30). For each test sample, the middle plots display the corresponding error of the predicted values, and the right plots show the differences between the real and predicted distributions of system parameters. The results indicate that the proposed DNN effectively extracts features from time series data and correlates them with system parameters, as the average error of these predictions is generally less than 5 percent.

\subsection{Lorenz system}
The Lorenz system, a typical example of a nonlinear dynamical system, is studied in this work. The system is governed by three equations, as shown below:
\begin{equation}\label{eq2}
\begin{split}
    \frac{dx}{dt}=\sigma(y-x) \\
    \frac{dy}{dt}=x(\rho-z)-y \\
    \frac{dz}{dt}=xy-\beta z
\end{split}
\end{equation}
where $\sigma, \rho$ and $\beta$ are system parameters, also referred to as simulation parameters in this work denoted by $P_{sim}=\left( \sigma, \rho, \beta \right)$. The Lorenz system exhibits chaotic behavior that is dependent on these three parameters, making it difficult to predict its dynamic behavior. However, in the following, it is demonstrated that the proposed data-driven method is capable of capturing the system's dynamics and identify system parameters.\par
The presented method's ability to identify system parameters of the Lorenz system is demonstrated in Figure \ref{lor1}. Specifically, Figure \ref{lor1}(a) examines the impact of changing parameter ${\sigma}$ by comparing predicted and actual values for 10 samples in the left plot. The middle plot displays the accuracy of the predicted system parameters as a percentage, while the right plot illustrates the predicted distribution versus the actual distribution for this parameter. The results demonstrate a strong overall performance, with only slight discrepancies between the predicted and actual distributions and average errors.\par
Identifying parameters accurately becomes more challenging when multiple parameters of a system are changed simultaneously. To illustrate this, we consider changing two parameters, $\sigma$ and $\rho$, of the Lorenz system and generating a corresponding dataset. The presented method is expected to identify both parameters and estimate their distributions.

Figure. \ref{lor2} presents the results of this experiment. As in Fig. \ref{lor1}, the performance is evaluated by comparing the predicted parameters with ground truth values for 10 samples, and by comparing the predicted and real distributions of the parameters. However, due to the added complexity of changing two parameters, the accuracy decreases for both parameters.

The left plot of Fig. \ref{lor2}(a) shows the comparison between the predicted and real values of $\sigma$ and $\rho$, while the middle plot shows the accuracy of the predicted system parameters as a percentage. The right plot displays the predicted and real distributions of $\sigma$ and $\rho$. As expected, the accuracy of the parameter estimation decreases compared to Fig. \ref{lor1}, as the framework has to identify both parameters simultaneously. Despite this, the overall performance is still satisfactory, as the discrepancies between the predicted and ground truth data remain relatively small.

\subsection{Flow passing over a cylinder}
Furthermore, our research presents a series of case studies that illustrate the efficacy of our proposed method for representing flow fields. We have designed our method to be applicable to a wide range of flows, without any restrictions on the type of flow considered. Nonetheless, we have specifically chosen to consider a two-dimensional flow field over a cylinder, which is a classic example commonly used in literature to demonstrate the usefulness of our approach \citep{raissi2020hidden}.

When the fluid passes through the cylinder, it generates a unique phenomenon known as the Karman vortex street, which is a pattern of vortices that form in the wake of the cylinder. This flow pattern is characterized by a steady-state flow with Reynolds numbers ranging between $R_{D}=100$ and $R_{D}=200$. The Navier-Stocks equations (NS) govern the behavior of the fluid flow and are described by the following equations:

\begin{equation} \label{navstok}
\begin{split}
&\nabla \cdot U=0 \\
&\frac{\partial U}{\partial t}=-\nabla \cdot (UU)-\nabla p+\frac{1}{R_{D}}\nabla^{2}U\\
\end{split}
\end{equation}
In these equations, $U$ represents the velocity of the fluid and $p$ represents the pressure. The stream-wise and transverse velocity components are represented by $u$ and $v$, respectively. The no-slip boundary condition is applied, and the channel is divided into a grid with 96 by 192 evenly spaced points for each sample. By analyzing the behavior of the flow field over the cylinder using our proposed method, we aim to showcase the effectiveness and versatility of our approach for studying various types of fluid flows.

In this section, we aim to quantify the Reynolds number of flows by analyzing their most important spatial and temporal features, and linking these features to the flow parameter known as the Reynolds number. To achieve this goal, we created time series of 2-D flow at different Reynolds numbers, ranging from 100 to 200, which encompassed the laminar regime of the flow. We divided these time series into training and test datasets.

We employed a combination of deep learning techniques, including a Convolutional Neural Network (CNN)-Autoencoder to extract compressed and relevant spatial features, a Long Short-Term Memory (LSTM)-Autoencoder to extract temporal features of the flow, and a Neural Network (NF) to interpret these features and determine the corresponding Reynolds numbers.

We collected two separate datasets for the velocity fields, one for the stream-wise velocity and one for the transverse velocity, with different Reynolds numbers. Figure. \ref{vel_v} demonstrates that the predicted Reynolds numbers for various Reynolds numbers for transverse velocity fields are very close to the real values. We also present the corresponding errors, as well as a comparison of the test dataset distribution and the prediction distribution in this figure.

The network has successfully extracted the spatial and temporal features and estimated the Reynolds number with NF, achieving a relatively low error rate. Similar performance was achieved for the streamwise velocity~(Fig. \ref{vel_u}), with an average error of less than 5 percent and the predicted distribution being very close to the real distribution.
\subsection{2-D transient cavity problem}
The objective of the lid-driven cavity problem is to gain a comprehensive understanding of the fluid's complex movements, which entail the formation of vortices and other flow patterns when the top lid is driven in a linear motion. This problem is classified as a transient phenomenon since it involves changes in fluid properties over time, in contrast to steady-state flow where the fluid properties remain constant. The Reynolds number in the training data ranges from 1000 to 6500. The channel is partitioned into 192 by 192 grids, and each sample set comprises 30 time steps. The behavior of fluid flow for the cavity problem is illustrated by the Navier-Stokes equations (NS) in equation. \eqref{navstok}.\par
The performance of the method presented for identifying the system parameter (Reynolds number) in a 2-D transient cavity problem is depicted in Fig. \ref{cavity}. As opposed to the flow over a cylinder case study, the accuracy decreases when dealing with transient conditions, where vorticity generates more complex patterns.
\subsection{Influence of Reynolds numbe on the prediction of parameters for flow case studies}
Our framework has a unique ability to identify dynamical systems. This means that it can analyze and understand the behavior of systems that change over time, such as fluid flow. In this section, we focus specifically on how the Reynolds number affects the predictions made by our DNN-NF model.

The Reynolds number is a dimensionless parameter that describes the flow of fluids. It is used to predict the transition from laminar flow to turbulent flow and is a crucial factor in understanding fluid mechanics. To analyze the impact of Reynolds number on our model's predictions, we trained our model on a dataset that included Reynolds numbers ranging from 400 to 3500 for stream-wise and transverse velocity over a cylinder, and between 1000 and 10000 for the cavity problem.It is worth noting that the channel grid remained the same for each case study. Additionally, each sample in the training set comprised 50 and 30 time steps for flow over a cylinder and cavity problem, respectively.

Figure. \ref{reynolds} presents the accuracy of the results for all three case studies as the Reynolds number increases. The results indicate that as the Reynolds number increases, the complexity of the flow also increases, making it more challenging for our model to accurately capture the most important features of the flow in both the spatial and temporal domains. Therefore, it is essential to consider the Reynolds number when using our model to predict fluid flow parameters, as it can have a significant impact on the accuracy of the results.

\section{Conclusion}
This study presents a DNN that uses LSTM autoencoder and Normalizing Flows together to determine the parameters of dynamic systems such as Duffing systems, Lorenz systems, and 2-D flow fields. We observed that by changing two parameters in Lorenz systems, the accuracy of predicted system parameters can decrease due to increased complexity in nonlinear system identification.
The spatial and temporal characteristics of the flow fields were successfully extracted through the use of CNN-autoencoders and LSTM-autoencoders, and were linked to Reynolds numbers of 2-D flows via NF. The results showed relatively low errors in predicted distributions, suggesting that the DNN architecture effectively captures the crucial features of the input data and accurately computes the relevant system parameters. We also observed that the accuracy of results for the 2-D cavity problem can decrease with increasing Reynolds number.

It is important to note that our proposed framework has certain potential limitations. Firstly, the range of nonlinearity and Reynolds numbers of interest is limited, which restricts the generalization of the method. Secondly, the accuracy of results decreases significantly when identifying multiple system parameters, as observed in the Lorenz system. Finally, future work should include the validation of our method on real-world dynamical systems, such as experimental case studies like fluid flows captured by Particle Image Velocimetry (PIV). Moreover, the integration of smart system design principles with nonlinear dynamics modeling supports innovative sustainability strategies, particularly by adapting dynamic solutions for environmental and structural applications \cite{shafa6ranking,shafa6smart}.

%
\section*{Conflict of interest}
The authors declare no conflict of interest.
\section*{Data Availability}
The data that support the findings of this study are available from the corresponding author upon reasonable request.
\section*{Funding}
This research was partially funded by the Physics of Artificial Intelligence Program of U.S. Defense Advanced Research Projects Agency (DARPA) and the Michigan Technological University faculty startup fund.
\bibliographystyle{unsrt}

\bibliography{my-collection}   

\end{document}